\documentclass[letterpaper, 10 pt, conference]{ieeeconf}  

\IEEEoverridecommandlockouts                              

\usepackage{microtype}
\usepackage{amsmath} 
\usepackage{amssymb}  
\usepackage{todonotes}
\let\labelindent\relax 
\usepackage{enumitem}
\usepackage{mathtools}
\usepackage{rotating}
\usepackage{multirow}
\usepackage{siunitx}
\usepackage{caption}
\usepackage{subcaption}
\usepackage{float}
\usepackage{hyperref}
\hypersetup{
    colorlinks=true,
    linkcolor=black,
    filecolor=magenta,      
    urlcolor=blue,
    pdftitle={Collective Perception Dataset Survey},
    pdfpagemode=FullScreen,
}
\usepackage{pifont}
\newcommand{\xmark}{\ding{55}}%
\usepackage{booktabs} 
\usepackage{graphicx}

\usepackage{algorithmic} 
\usepackage{algorithm2e}
\graphicspath{{graphics/}}

\usepackage{tabto} 

\title{\LARGE \bf
Collective Perception Datasets for Autonomous Driving:\\ A Comprehensive Review 
}

\author{Sven Teufel$^{1}$, Jörg Gamerdinger$^{1}$,  Jan-Patrick Kirchner$^{1}$, Georg Volk$^{1}$, and Oliver Bringmann$^{1}$
\thanks{$^{1}$University of T\"ubingen, Faculty of Science, Department of Computer Science, Embedded Systems
\tt\small {\{sven.teufel, joerg.gamerdinger, jan-patrick.kirchner,  georg.volk,  oliver.bringmann\} @uni-tuebingen.de}}
}%

\begin{document}
\maketitle
\thispagestyle{empty}
\pagestyle{empty}

\begin{abstract}
To ensure safe operation of autonomous vehicles in complex urban environments, complete perception of the environment is necessary. However, due to environmental conditions, sensor limitations, and occlusions, this is not always possible from a single point of view. To address this issue, collective perception is an effective method. Realistic and large-scale datasets are essential for training and evaluating collective perception methods.
This paper provides the first comprehensive technical review of collective perception datasets in the context of autonomous driving. The survey analyzes existing V2V and V2X datasets, categorizing them based on different criteria such as sensor modalities, environmental conditions, and scenario variety. The focus is on their applicability for the development of connected automated vehicles. This study aims to identify the key criteria of all datasets and to present their strengths, weaknesses, and anomalies. Finally, this survey concludes by making recommendations regarding which dataset is most suitable for collective 3D object detection, tracking, and semantic segmentation.
\end{abstract}
\section{INTRODUCTION}
\label{sec:intro}
According to a report by the European Union in 2019, human error is responsible for \SI{90}{\percent} of fatalities in traffic accidents~\cite{europeanUnion2019}. Autonomous vehicles have the potential to reduce traffic fatalities and improve traffic flow. However, accurate perception of the environment is crucial for their success. Vehicle-local perception faces several challenges, including occlusions and limited sensing ranges, which can hinder environmental perception. Exchanging information about the environment through Vehicle-to-Vehicle (V2V) or Vehicle-to-Everything (V2X) communication enables the use of data from distributed connected automated vehicles (CAVs) to address this problem ~\cite{volk-2019-environment-aware, gamerdinger-2023-cold}.

For vehicle-local perception many datasets such as KITTI~\cite{kitti}, Waymo~\cite{waymo}, and nuScenes~\cite{nuScenes} exist which are widely used in research. For the development and evaluation of algorithms for collective perception, these datasets cannot be used because they only contain sensor data from a single point of view. Therefore, in recent years, various datasets for collective perception have been published, each with their own advantages and disadvantages.  In this paper, we provide a meticulous analysis of these publicly available datasets to gain deep insights. We aim to objectively present the advantages and limitations of different datasets to recommend the most suitable one for a given task. 

The comprehensive dataset review with the focus on V2V and V2X collective perception datasets is shown in Section~\ref{sec:review}. We separated the review in two parts. Section~\ref{sec:simulated} shows the simulated collective perception datasets. First fundamental information about the tooling, used to generate the datasets, are introduced, then we present an in-depth analysis of the simulated V2V and V2X datasets, as well as an overview on infrastructure-only collective perception datasets. Section~\ref{sec:real} gives a detailed presentation of the real-world V2V and V2X datasets and an overview on infrastructure-only datasets.
Afterwards, we discuss the results of the analysis in Sec.~\ref{sec:discussion}. Finally in Sec.~\ref{sec:conclusion} we conclude our review.

\begin{figure}[t]
    \centering
    \includegraphics[width=0.95\linewidth, clip]{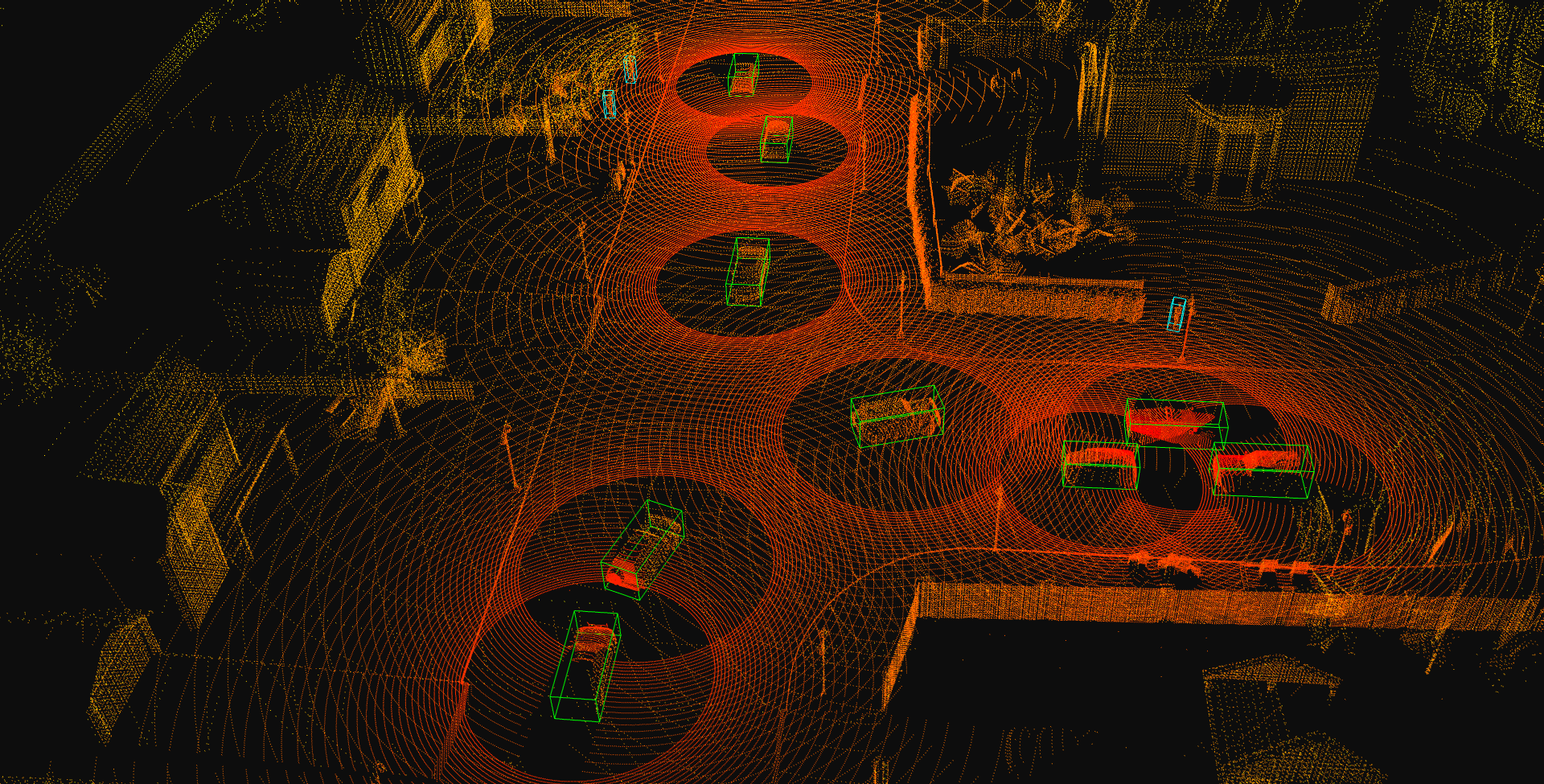}
    \caption{Collective LiDAR point cloud from CODD ~\cite{codd}} 
    \label{fig:lateral}
\end{figure}
\section{DATASET REVIEW}
In this section we provide details about the available datasets. A detailed overview of V2V and V2X collective perception datasets is given in Tab. \ref{tab:overview}. For some datasets information, such as sensor setup or scenarios, are missing as no analysis was possible due the fact that the dataset is not publicly available and the information were not provided by the authors.
\label{sec:review}
\subsection{Simulated Datasets}
\label{sec:simulated}
Collecting a real-world dataset leads to several problems, as many scenarios cannot be recorded due to high risk. Furthermore environmental influences such as rain or snow are not reproducible and equipping multiple test vehicles with sensors would be too costly and time consuming. Thus, most collective perception datasets are synthetic datasets generated using simulators such as the open source simulator CARLA~\cite{CARLA}. CARLA allows to simulate urban, sub-urban, rural, and highway scenarios on a variety of maps, including weather conditions, with realistic rendering. Vehicles can be equipped with different sensors like RGB cameras, depth cameras, radar and LiDAR. However, the LiDAR sensor integrated in CARLA does not have physically correct dropout and intensity calculation. The traffic management can be performed by CARLA or in a co-simulation with SUMO~\cite{sumo}. SUMO is a road traffic simulator that allows to generate specific routes for different classes of vehicles and vulnerable road users (VRUs), which can then be simulated in CARLA. Xu et al.~\cite{xu2021opencda} proposed OpenCDA, a framework for prototyping collective perception. The framework integrates a CARLA and SUMO co-simulation and allows scenario management and benchmarking for perception and planning.
\vspace*{0.2cm}
\subsubsection{\textbf{V2V-Sim}}
The V2V-Sim dataset was released in 2020 by Wang et al.~\cite{wang2020v2vnet}. In contrast to all other synthetic datasets, they used LiDARsim~\cite{manivasagam2020lidarsim} instead of CARLA to generate their recordings. LiDARsim allows to build maps and insert vehicle meshes from a large database. After performing a raycasting a special trained neural network performs the raydrop to achieve realistic point clouds. In V2V-Sim, scenarios from the real-world ATG4D dataset~\cite{yang2018pixor} were rebuild in the simulator to achieve realistic traffic. On average 10 CAVs are present per sample, this number can increase up to 63. In total they generated 5,500 snippets with \SI{25}{\second} each, resulting in 46,796 frames for training and 4,404 frames for testing. The main task for this dataset is collective 3D object detection.
\vspace*{0.2cm}
\subsubsection{\textbf{CODD}}
Arnold et al.~\cite{codd} presented the CODD dataset in 2021. CODD is a simulated dataset for collective perception and SLAM that contains only LiDAR point clouds. This dataset was generated using CARLA as a simulator. In total there are 108 scenarios with 4 to 16 CAVs in urban, sub-urban and rural environments, resulting in 13,500 frames. As additional traffic only pedestrians but no non-cooperative vehicles are present in the dataset. A 64-layer \SI{360}{\degree} LiDAR sensor with a maximum range of \SI{100}{\meter} was used to record the point clouds. The recording frequency of the LiDAR sensor was \SI{5}{\hertz}, resulting in a relatively low rate of 50,000 points per second. The main task for this dataset is collective 3D object detection, for which 204,250 3D bounding boxes in the car and pedestrian classes are provided. This dataset can be used for training and evaluation of point cloud registration methods, for which a benchmark is also provided with 6,129/1,339/1,315 pairs of point clouds for training, validation and testing. Furthermore, this dataset can be used for object tracking and SLAM. Since there are no camera images included, this dataset cannot be used for training or evaluation of camera-based methods or LiDAR-camera fusion methods. However, it is well suited for LiDAR-only methods for 3D object detection, especially the relatively high number of CAVs and the presence of pedestrians, which allows to study the impact of different CAV rates on the perception performance as well as the perception of VRUs. The relatively low recording rate of \SI{5}{\hertz} leads to larger position offsets between frames, which could negatively affect multi-frame detection or tracking methods. The code used to generate this dataset is publicly available, so this dataset can be easily reproduced and extended.
\vspace*{0.2cm}
\subsubsection{\textbf{COMAP}}
The COMAP dataset by Yuan et al.~\cite{comap} was presented in 2021. COMAP is a simulated collective perception dataset generated with a co-simulation of CARLA and SUMO. This dataset consists of 21 intersection scenarios on CARLA's \textit{Town05} map. In each scenario there are between 2 and 20 CAVs, each equipped with a 32-layer \SI{360}{\degree} LiDAR sensor, recording 200,000 points per second with a range of \SI{80}{\metre} and a RGB camera with a resolution of 800$\times$600\,px, both recording at a frequency of \SI{10}{\hertz}. In total, COMAP consists of 8,668 frames, 4,655 for training and 4,013 for testing in the provided benchmark. Besides the CAVs, there are several other non-cooperative vehicles as traffic in each scenario, other road users such as VRUs are not included. In total, there are 226,958 annotated 3D bounding boxes of vehicles (in the classes of passenger car, electric vehicle, truck, authority vehicle and cooperative vehicle), as well as semantic labels for the LiDAR point clouds and camera images. The main tasks for this dataset are collective 3D object detection and semantic segmentation for LiDAR and camera. As this dataset only includes vehicles, it is not appropriate for evaluating detection methods for other road users. Additionally, it consists only of similar intersection scenarios, and is not diverse, which may lead to overfitting of the training data, and since the training and test data are very similar, the obtained results may not be representative. However, due to the wide range in the number of CAVs, this dataset can be used to investigate the impact of different CAV rates on perceptual performance. The code which was used to generate this dataset is publicly available, therefore this dataset can be easily reproduced and extended.
\vspace*{0.15cm}
\subsubsection{\textbf{V2XSet}} 
V2XSet was presented by Xu et al.~\cite{xu2022v2xvit} in 2022. CARLA and OpenCDA were used for the data generation. With a train/validation/test split of 6,694/1,920/2,833 frames, V2XSet has 11,447 frames in total with 254,582 labeled vehicles. These frames are distributed among 55 scenarios with five varying road types on 8 different maps of CARLA. The most frequent scenarios are intersections with about \SI{50}{\percent}. There are between 2 and 7 collaborative agents in each scenario, each equipped with a 32-layer \SI{360}{\degree} LiDAR sensor at the top with \SI{120}{\metre} range and a frequency of \SI{10}{\hertz}. In addition, the vehicles are equipped with four RGB cameras (800$\times$600\,px) providing a \SI{360}{\degree} surround view. Besides the CAVs there is up to one road side unit (RSU) in the intersection, mid-block and entrance ramp scenario.
The main task for this dataset is collective 3D object detection.
An advantage of the dataset is the realistic noise simulation in the communication and the variety of scenarios which allow a comprehensive evaluation. 
During the analysis of the dataset we found scenarios in which a collision occurred.
Due to the collision the vehicle mesh does not match to the ground truth bounding box for several frames. Moreover, the dataset only contains vehicles and no VRUs which limits the applicability. Furthermore, the positioning of the vehicles in one highway scenario can be seen as anomaly or at least as unrealistic, as there are multiple groups of vehicles parallel on all four lanes with only about \SIrange{10}{15}{\metre} distance between the groups.
\vspace*{0.2cm}
\subsubsection{\textbf{DOLPHINS}}
The DOLPHINS dataset was presented by Mao et al.~\cite{dolphins} in 2022. The dataset was generated using the CARLA simulator and consists of six scenarios. Four out of the six scenarios are intersection or T-junction scenarios. Additionally, the dataset contains a highway scenario with a merging lane and one scenario on a rural road. Each scenario consists of about 7,000 frames leading to 42,376 frames and 292,549 objects, with 2D as well as 3D annotations in KITTI format. Vehicles are labeled in a range of 200$\times$80\si{\metre} around the ego vehicle. Dolphins allows not only V2V collective perception but also V2X by including a RSU. Besides the ego there are two further viewpoints (two RSUs, RSU+CAV or two CAVs) in each scenario. The vehicles and the RSUs are equipped with a high resolution RGB camera (1920$\times$1080\,px) and a 64-layer \SI{360}{\degree} LiDAR sensor with a measurement range of \SI{200}{\metre} and 2.56M points per second at a rotation frequency of \SI{20}{\hertz}; however, the sensor data was only captured at a frequency of \SI{2}{\hertz}. Additionally to the CAVs, all scenarios contain a varying number of non-cooperative vehicles and pedestrians. The main task for this dataset is collective 3D object detection, for which also a benchmark with an 5:3 train/test split is provided. In this dataset the number of viewpoints is relatively low in all scenarios. An advantage of this dataset are the rain and fog effects in two scenarios, which allow to evaluate perception algorithms under adverse weather; however, only for image-based detection, since the LiDAR in CARLA is not affected by weather effects. As the code used for dataset generation is publicly available, the dataset is reproducible and can be extended easily.
\begin{figure}[t]
    \centering
    \includegraphics[width=\linewidth, trim= 3.5cm 2cm 3.5cm 2cm, clip]{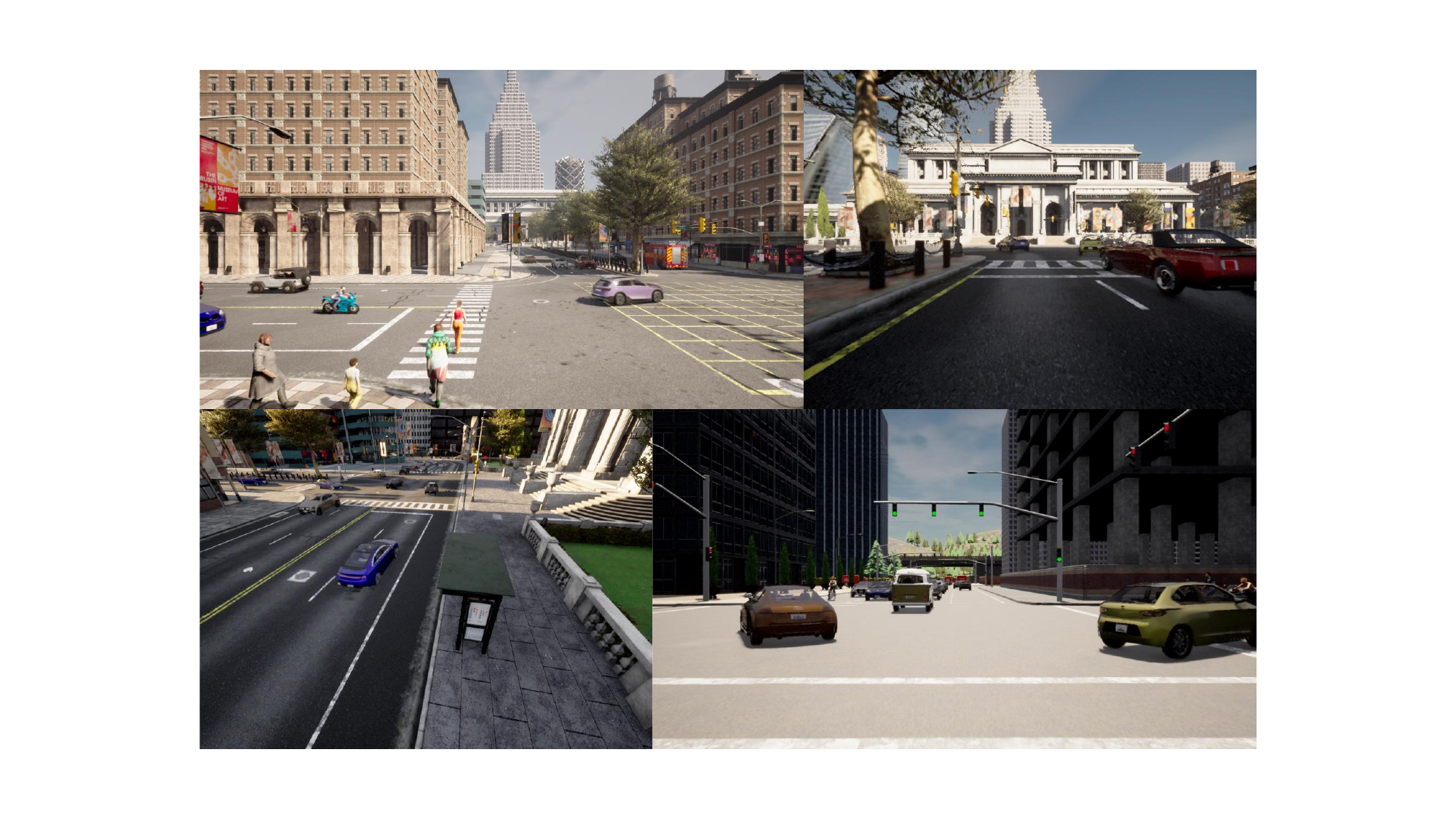}
    \caption{Exemplary scenes from simulated datasets with images. DOLPHINS (top left), OPV2V (top right), V2XSet (bottom left), and V2X-Sim (bottom right)} 
    \label{fig:result_showcase}
\end{figure}
\vspace*{0.2cm}
\subsubsection{\textbf{OPV2V}}
OPV2V was presented by Runsheng et al.~\cite{opv2v} in 2022. OPV2V was generated using the OpenCDA co-simulation with CARLA and SUMO. This dataset consists of 73 scenarios in urban, sub-urban, rural, and highway environments. The scenarios were collected in 8 CARLA towns and in a digital model of Culver City. In total, OPV2V consists of 11,464 frames with 232,913 annotated 3D vehicle bounding boxes.  There are between 2 and 7 CAVs in each scenario, with an average of approximately 3 CAVs, each equipped with a 64-layer \SI{360}{\degree} LiDAR sensor with \SI{120}{\meter} range, and 1.3M points per second. In addition, each CAV is equipped with 4 RGB cameras facing left, right, front and back with a resolution of 800$\times$600\,px, furthermore the CAVs were equipped with GPS and IMU. Data from all sensors is recorded at \SI{10}{\hertz}. Besides the CAVs, there are several other non-cooperative cars as traffic in each scenario, further types of road users are not included. The 3D bounding boxes are stored in separate files for each CAV and contain only the currently visible objects, i.e. there must be at least one LiDAR point on the object for the label to be included. In order to get all the labels in one frame, the labels of all the CAVs must be merged. In addition to the annotated 3D boxes, the dataset includes semantic bird's-eye view (BEV) maps for each CAV, as well as the speed and planned trajectory for each CAV. The main task for this dataset is collective 3D object detection, for which a benchmark with a split of 6,764/1,981/2,719 frames for training, validation and testing is provided. It can also be used for BEV semantic segmentation, tracking and prediction. OPV2V is well suited for training and evaluation of collective 3D object detection methods for LiDAR and camera due to its large scale, the provided benchmark and the variety of scenarios, however the number of CAVs in each scenario is relatively low (average $\approx 3$) and the cameras have a low resolution (800x600px), furthermore there are no road users other than cars, so the detection of other road users cannot be evaluated. The code used to generate this dataset is publicly available, so this dataset can be easily reproduced and extended.
\vspace*{0.2cm}
\subsubsection{\textbf{V2X-Sim}}
V2X-Sim Version 1.0 was released in 2021 by Li et al.~\cite{v2xsimv1}. V2X-Sim V1.0 was generated using a co-simulation of CARLA and SUMO, containing 100 intersection scenarios on CARLA map $Town05$ with 100 frames each, resulting in 10,000 frames in total. The dataset is stored in the same format as the nuScenes \cite{nuScenes} dataset. In each scenario 2 to 5 vehicles were selected as CAVs, each equipped with a 32-layer \SI{360}{\degree} LiDAR sensor, with a measurement range of \SI{70}{\metre} and a frequency of \SI{20}{\hertz}, however the sensor data was recorded at only \SI{5}{\hertz}. The CAVs were also equipped with a RGB camera, but the camera images are currently not included in the dataset. The main task for this dataset is LiDAR-based 3D object detection, for which also a benchmark with 8,000/900/1,100 frames for training/validation/testing is provided. Furthermore, the low variation in scenarios may lead to overfitting and the results might not be representative due to the similarity in the training and test data. 

The second version of V2X-Sim ~\cite{V2XSim} was released in 2022, it extends the first version with more modalities, scenarios and tasks. The CAVs are equipped with the same 32-layer LiDAR as in Version 1.0, but there are additional 6  high resolution RGB cameras (1600$\times$900\,px) positioned around the vehicle to get a \SI{360}{\degree} camera coverage. Additionally to the CAVs there were also RSUs added at the intersections, each equipped with the same LiDAR as the CAVs and 4 RGB cameras (1600$\times$900\,px). V2X-Sim Version 2.0 consists also of 100 intersection scenarios in urban and sub-urban environments with 100 frames each, but on more CARLA maps, namely $Town03$, $Town04$ and $Town05$. The number of CAVs is again randomly selected between 2 and 5. The provided labels are not only 3D bounding boxes in the classes car, cyclist, motorcyclist, but also semantic labels for LiDAR and camera, BEV semantic labels and depth labels for the cameras. The main tasks, for which benchmarks are also provided are collective BEV object detection, BEV semantic segmentation and BEV tracking with a 8,000/1,000/1,000 split for training/validation/testing. The broad amount of different labels and the provided benchmarks make this dataset suitable for a wide range of tasks, however the diversity in scenarios, sensors and the number of CAVs is still limited. 

\vspace*{0.2cm}
\subsubsection{\textbf{IRV2V}}
IRV2V was published as the first collective perception dataset with different temporal asynchronies in 2023 by Wei et al.~\cite{wei2023asynchrony}. The authors use the CARLA simulator to generate the dataset with 2 to 5 CAVs per scene. The CAVs are equipped with 4 RGB cameras with a resolution of 600$\times$800\,px and an 32-layer \SI{360}{\degree} LiDAR sensor. In total, the dataset consists of 33,796 images and 8,449 point clouds. For collective perception this can be considered as 8,449 frames with a frequency of \SI{10}{\hertz} which are split into 5,445 frames for training, 2,010 frames for validation and 994 for testing. With an average of about 48 and up to 113 vehicles per scene, there is a total of 1,564,033 vehicles present in the dataset from which 1,203,793 are moving vehicles. 
The main task for the dataset is collective bird's eye view perception under time asynchronies.
\vspace*{0.2cm}
\subsubsection{\textbf{DeepAccident}}
Wang et al.~\cite{wang2023deepaccident} published the DeepAccident dataset in 2023. This is a synthetic V2X dataset generated using the CARLA simulator; it contains 12 types of accident scenarios frequently occurring in real-world. Therefore, different places such as straight roads, rural roads, highways and (un)signalized intersections are considered under various weather and different lighting conditions. Moreover, the number of further surrounding road users varies. Each scenario consists of an ego, three CAVs and one infrastructure sensor set. The ego and one CAV are colliding during the scenario, the other CAVs are following the colliding vehicles. The four vehicles and the RSU are equipped with 6 RGB cameras with 1600$\times$900\,px and \SI{70}{\degree} FoV providing a \SI{360}{\degree} surround view and one 32-layer \SI{360}{\degree} LiDAR sensor with \SI{70}{\metre} measurement range, recording at \SI{10}{\hertz}. The dataset consists of 691 scenarios with 57k frames and 285k annotated samples (trajectories and 3D bounding boxes) which are split according to the scenario split into 203k/41k/41k samples for train, validation and test. These annotated objects include six classes which are car, van, truck, motorcycle, cyclist, and pedestrian.
The main task of the DeepAccident dataset is collective motion and accident prediction, but also collective 3D object detection and tracking is possible. Due to the high number of frames and the variance of scenarios the dataset is suitable to train and evaluate collective 3D object detection. By including different times of day and weather conditions, also an evaluation of camera-based object detection under adverse weather is enabled. Moreover, the dataset covers the most relevant object classes and includes VRUs. DeepAccident is the only collective perception dataset focusing on accidents and therefore, is the most suitable for motion and accident prediction. However, due to the consistent number of collaborative agents, no varying V2X equipment rates can be investigated for collective 3D object detection.
\vspace*{0.2cm}
\subsubsection{\textbf{Coop3DInf}} 
The Coop3DInf dataset was published by Arnold et al.~\cite{arnold2020cooperative} in 2020 and is a synthetic infrastructure-only dataset generated using the CARLA simulator.
The data is collected using RGB and depth cameras with a resolution of 400$\times$300\,px, \SI{90}{\degree} FoV and a frequency of \SI{10}{\hertz}. The depth image is used to generate 3D data points to form a point cloud similar to a LiDAR sensor, but without an intensity value. The dataset consists of a T-junction and a roundabout scenario, which are used to collect two independent image collections for each scenario, one for training with 4,000 frames and one for testing with 1,000 frames, leading to a total of 10,000 frames. The dataset contains 121,225 labeled objects with three object classes: vehicles, cyclists/motorcyclists (not distinguished), and pedestrians. The main task is collective 3D object detection using infrastructural sensors; however, due to the insufficient sensor setup with a calculated point cloud from an image the applicability and realism is limited.


\begin{figure*}
    \centering
    \includegraphics[trim=0 2.5cm 0 2cm ,clip, width=1\linewidth]{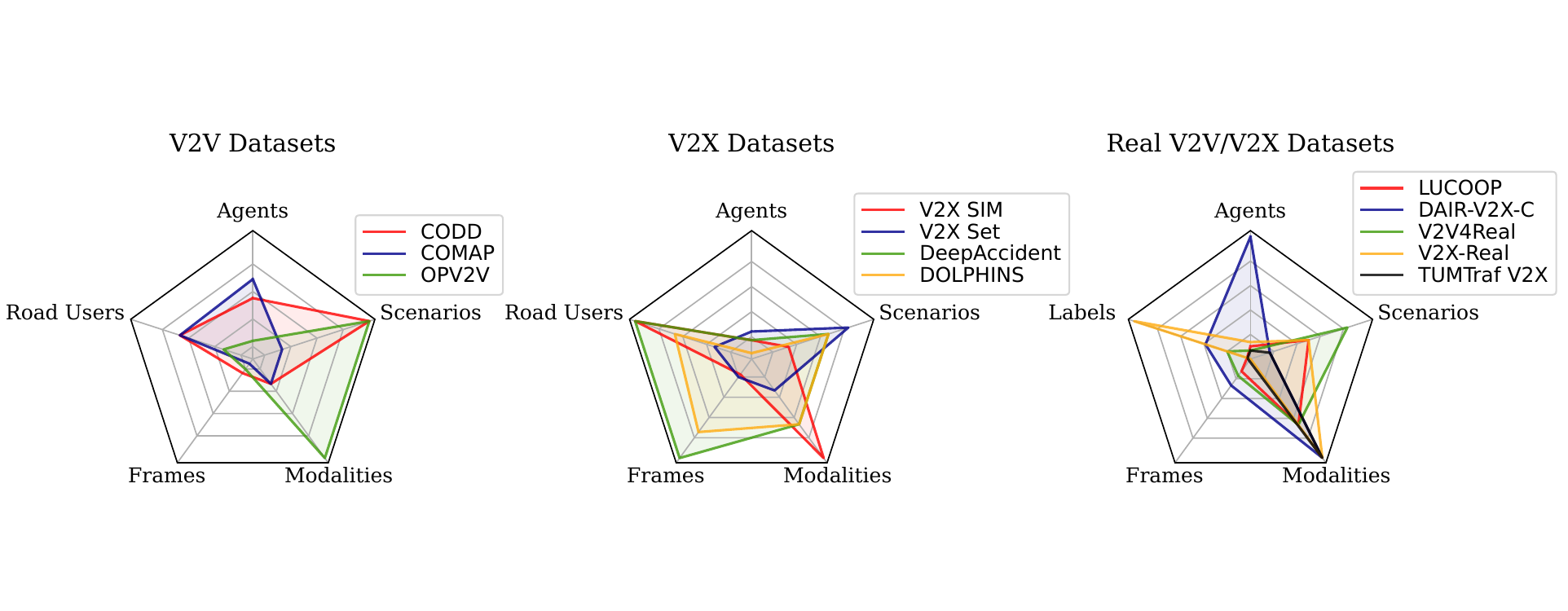}
    \caption{Overview of the presented V2V and V2X datasets. Datasets where information are missing and infrastructure-only datasets are not shown here, an overview on all V2V and V2X datasets is shown in Tab. \ref{tab:overview}.}
    \label{fig:radar}
\end{figure*}

\subsection{Real-World Datasets:}
\label{sec:real}
\subsubsection{\textbf{T\&J}}
The T\&J dataset by Chen et al.~\cite{cooper, f-cooper} was the first V2V collective perception dataset and was released in 2019. They equipped two golf carts with a 16-layer \SI{360}{\degree} LiDAR sensor, two 2,048$\times$1,544\,px cameras, radar, GPS sensor and an IMU. The available dataset consists of 100 frames from the two golf carts on a parking lot. The main task of the dataset is collective 3D object detection; however, there are no labels for the objects available which makes a quantitative evaluation impossible. Also for training neural networks the dataset is not suitable due to the missing labels and the low number of frames. Moreover, they do not provide any details about the number of available objects or images.
\vspace*{0.2cm}
\subsubsection{\textbf{DAIR-V2X}}
DAIR-V2X was presented in 2022 by Yu et al. \cite{dair-v2x} and was the first large scale real-world dataset for collective perception. DAIR-V2X was recorded in Beijing, China and contains data from multiple infrastructure sensors at intersections as well as from one CAV.  In total, DAIR-V2X contains data from 28 intersections, each equipped with two 300-layer \SI{100}{\degree} LiDAR sensors with a maximum range of \SI{280}{\metre} and with two 1920$\times$1080\,px RGB Cameras. The sensor data is recorded at \SI{10}{\hertz} and \SI{25}{\hertz} for the LiDAR sensors and cameras respectively. The CAV was equipped with a 40-layer \SI{360}{\degree} LiDAR and one 1920$\times$1080\,px RGB camera facing to the front of the vehicle. Additionally, the CAV was equipped with an IMU and GPS. DAIR-V2X is composed of 3 subsets, which contain different parts of the collected data. 

DAIR-V2X-C contains both recordings from the CAV and the RSUs, resulting in 13,000 frames in total. The data from 100 measurement drives through the sensor equipped intersections is sampled at \SI{10}{\hertz} and contains the CAVs frames as well as the closest frames in time of the RSU. Since the vehicle sensors are not synchronized with the infrastructure sensors the time difference between the CAV and RSU frames varies. The other subsets are DAIR-V2X-V, which only contains recordings from the CAV and DAIR-V2X-I, which only contains recordings from the RSUs. Since DAIR-V2X-V and DAIR-V2X-I do not contain collective information, i.e. multiple sensor views of the same scene, we focus only on the DAIR-V2X-C dataset here. DAIR-V2X-C includes 464k 3D labels in the classes car, truck, van, bus, tricyclist, motorcyclist, cyclist, pedestrian and traffic cone. Trajectories or information of the vehicle dynamics are not included. The main task for this dataset is collective 3D object detection.  For this purpose, a benchmark for camera and LiDAR at different fusion stages with a train/validation/test split of 5:2:3 is provided. Since this dataset features a huge amount of infrastructure sensors, it is especially suitable to study V2I collective perception. Since there is only one CAV, it is not suitable for V2V collective perception.
\vspace*{0.2cm}
\subsubsection{\textbf{V2X-Seq}} Based upon the DAIR-V2X dataset, Yu et al. published the V2X-Seq dataset \cite{v2xseq} in 2023. V2X-Seq consists of two parts, the sequential perception and the trajectory forecasting. The sequential perception dataset is comprised of 15k frames from 95 scenarios with a length of \si{10}-\SI{20}{\second}. The scenarios are the same as in DAIR-V2X-C, the main difference to DAIR-V2X-C is, that in V2X-Seq sequential perception, each object is labeled with a unique tracking id, so that the trajectory of moving objects can be tracked. The main task for this dataset is collective object tracking. While the sequential perception dataset can also be used for trajectory forecasting, it is quite small for this task, this is why Yu et al. also published the trajectory forecasting dataset. The trajectory forecasting dataset consists of selected trajectories from \SI{336}{\hour} of recorded data from the CAV an the RSUs each. In total this dataset is comprised of 50,000 V2I scenarios, as well as 80,000 vehicle and infrastructure-only scenarios. Each of these scenarios includes the tracking information of objects in 8 classes for \SI{10}{\second} at a frequency of \SI{10}{\hertz}. Additionally to the labeled objects, traffic light information as well as maps of the 28 intersections are included. As the name implies the main task for this dataset is the prediction of the future trajectory of objects using V2I information. V2X-Seq provides benchmarks for collective tracking and trajectory forecasting with the same train/validation/test split ratios as in DAIR-V2X-C on its corresponding subsets, it is well suited to study these tasks for V2I collective perception.
\vspace*{0.2cm}
\subsubsection{\textbf{V2V4Real}}
The V2V4Real dataset was presented in 2023 by Xu et al.~\cite{v2v4real}. The V2V4Real dataset consists of 10k frames recorded with \SI{10}{\hertz} on \SI{347}{\kilo\metre} highway and \SI{63}{\kilo\metre} urban roads including intersections, straight roads and highway entrance ramps. For recording two vehicles equipped with a mono camera to the front and back (CAV 1: 2064$\times$1544\,px, CAV 2: 1280$\times$720\,px) and a \SI{360}{\degree} 32-layer LiDAR with \SI{200}{\metre} measurement range. Additionally, the vehicles are equipped with a GPS/IMU systems. The main task for this dataset is collective 3D object detection, for which the dataset provides 240k 3D bounding boxes including the driving state of five object classes (car, pickup truck, van, semi-truck, bus). Additionally, our analysis has shown that the dataset also contains pedestrians which are not mentioned in the paper. Bounding boxes are labeled in the corresponding relative coordinate system of the detecting LiDAR sensor which also allows local detection evaluation. The dataset is not only applicable for detection but also for tracking, however, the number of CAVs is very low in this dataset. Moreover, about \SI{80}{\percent} of the dataset is recorded on highways which normally show less occlusions and necessity for collective perception.
\vspace*{0.15cm}
\subsubsection{\textbf{LUCOOP}}
LUCOOP was published by Axmann et al. \cite{lucoop} in 2023. LUCOOP is a real-world dataset which was recorded in the city of Hannover with three CAVs. Each CAV was equipped with a different sensor setting, on the first CAV there were two LiDAR Sensors mounted, one 64-layer \SI{360}{\degree} LiDAR mounted horizontally and one 16-layer \SI{360}{\degree} LiDAR mounted vertically at the back of the CAV. The second and third CAV were equipped with a 32-layer \SI{360}{\degree} LiDAR and a 16-layer \SI{360}{\degree} LiDAR respectively. All LiDARs recorded at a frequency of \SI{10}{\hertz}. Additionally, all CAVs were equipped with GNSS, IMU and UWB range measurement sensors. The UWBs were used to determine the distance between the CAVs as well as to a station along the driving route. Furthermore to accurately measure the position of CAV 1 a total station was used along the route. LUCOOP consists of three CAVs driving in convoy on an inner city round course. CAV 1 and 2 completed four rounds, while CAV 3 completed two rounds. The dataset comprises 15k frames for CAV 1 and 2, and 7k frames for CAV 3. Additionally, 7k merged point cloud frames from all three CAVs are provided. Besides the trajectories of all three CAVs, also 3D bounding boxes of the surrounding objects in the classes car, bicycle, pedestrian, truck and parked car are provided, however only the merged point clouds are labeled, resulting in 7k labeled frames with 34k labeled objects. The main tasks for this dataset are urban navigation, including localization, path planning and SLAM as well as 3D object detection. Due to the highly accurate ground truth for the vehicle trajectories, this dataset is well suited for the navigation related tasks. For the object detection task this dataset is not that well suited, since the variation is very low and the same scenes are recorded multiple times, which can lead to overfitting, however, the heterogeneous sensor setups of the CAVs allows to study the domain adaptation and early fusion for different sensor types. Also it is the only real-world dataset featuring 3 CAVs to this date. 
\vspace*{0.15cm}
\subsubsection{\textbf{Ko-Per}}
The 2014 released Ko-PER intersection laser scanner and video dataset by Strigel et al.~\cite{ko-per2014} was the first collective perception dataset, but an infrastructure-only dataset. The dataset consists of one four-way intersection scenario in Aschaffenburg, Germany. The intersection is equipped with 14 8-layer solid-state LiDARs and 8 monochrome cameras with 656$\times$494\,px; but, only two out of eight camera images are provided. The dataset contains 3D bounding boxes with cars, trucks, pedestrians, and cyclists in 4,850 frames in total. 3D collective object detection can be seen as main task for the dataset.
\vspace*{0.15cm}
\subsubsection{\textbf{Cityflow}}
In 2019, Tang et al.~\cite{tang2019cityflow} published the Cityflow dataset, which is an infrastructure-only image dataset. It consists of \SI{3}{\hour} synchronized videos from 40 cameras (1920$\times$1080\,px) which are distributed at 10 intersections over \SI{2.5}{\kilo\metre}. The dataset covers varying traffic densities in highway and residential areas. In total, the dataset consist of 666 vehicles which are annotated with 229,680 2D bounding boxes. The main task is multi-target multi-camera tracking, thus only vehicles passing at least two cameras are labeled.
\vspace*{0.15cm}
\subsubsection{\textbf{IPS300+}}
The IPS multimodal dataset for intersection perception systems by Wang et al.~\cite{wang2021ips300} was released in 2021. The dataset was recorded at a single intersection in Beijing, China. They use 2 sensor units with 2 RGB cameras with a 1920$\times$1080\,px resolution and one \SI{360}{\degree} LiDAR sensor with 80 layers and a measurement range of \SI{230}{\metre}. With an annotation frequency of \SI{5}{\hertz} they provide 14,198 frames in total. Special for this dataset is the high number of 3D labels with an average of about 320 objects per frame which are distributed over the classes car, cyclist, pedestrian, tricyclist, bus, truck, and engineer car. Collective 3D object detection is the main task.
\vspace*{0.15cm}
\subsubsection{\textbf{WIBAM}}
The WIBAM dataset was presented by Howe et al. \cite{wibam} in 2021 and is an infrastructure camera-only dataset. WIBAM was recorded with 4 unsynchronized RGB cameras with 2560$\times$1440\,px at \SI{50}{\hertz}. After synchronization the images were sampled at \SI{12.5}{\hertz} with a resolution of 1920$\times$1080\,px. In total, WIBAM contains 33k images, which are split into three subsets for training, validation and testing with a size of \SI{75}{\percent}, \SI{15}{\percent} and \SI{10}{\percent} respectively. The train and validation set contain 116,702 automatically generated 2D bounding boxes, while the test set contains 1,651 manually labeled 3D bounding boxes. The main task for this dataset is 2D/3D object detection.
\subsubsection{\textbf{LUMPI}}
The LUMPI infrastructure-sensor perception dataset by Busch et al.~\cite{busch2022lumpi} was released in 2022. The dataset was recorded on one large intersection in Hannover, Germany. For recording up to 3 RGB cameras were used, one with 1640$\times$1232\,px and the others with 1920$\times$1080\,px. Moreover, the recordings consist of data from up to five different \SI{360}{\degree} LiDAR sensors with 16 to 64 layers. The sensors reach measurement ranges from \SIrange{20}{200}{\metre}. In total they captured seven measurements with a recording time of \SI{145}{\minute} consisting of 200k images and 90k point clouds. The main task is collective 3D detection; however, there are no labels available yet.
\subsubsection{\textbf{TUMTraf A9 Highway}}
The TUMTraf A9 Highway dataset (former A9 dataset) was published by Creß et al. in 2022 \cite{cress2022a9}. TUMTraf is an infrastructure-only dataset, that was recorded on a \SI{3}{\kilo\metre} long segment of the german highway A9 near Munich. This dataset contains data from 1920$\times$1200\,px RGB cameras and 64-layer \SI{360}{\degree} LiDAR sensors with \SI{120}{\metre} measurement range, mounted at two gantry bridges over the highway. It is structured into three subsets, the first subset contains 600 random camera images from four cameras. The second subset contains a \SI{25}{\second} sequence of camera images at \SI{2.5}{\hertz}. The third subset consists of two sequences of LiDAR point clouds, recorded at \SI{2.5}{\hertz}. All subsets are labeled with 3D bounding boxes, the second subset additionally contains unique ids for the labeled objects. In total, this dataset consists of 1000 frames and 14k labeled objects. In the second release TUMTraf was extended by 3 scenarios that were recorded by four synchronized RGB cameras at \SI{10}{\hertz}, it includes a snowy, foggy and accident scenario. The scenarios are labeled with 3D bounding boxes and unique object ids.

\subsubsection{\textbf{TUMTraf Intersection}}
The TUMTraf Intersection dataset (former A9 Intersection dataset) was presented by Zimmer et al. \cite{zimmer2023a9} in 2023. It contains recordings from two LiDAR sensors and two cameras mounted at an intersection gantry bridge near the city of Munich. The used sensors were the same as in the TUMTraf A9 Highway dataset and recorded at a frequency of \SI{10}{\hertz}. In total this dataset contains 4.8k time synchronized frames. This dataset provides 57.4k labeled 3D bounding boxes in 10 classes as well as unique ids for all objects. Besides recordings under good weather conditions during the day, also one scenario at night time under rainy conditions is included. The main task for this dataset is collective 3D object detection.

\vspace*{0.2cm}
\subsubsection{\textbf{TUMTraf V2X}}
The TUMTraf V2X Collective Perception dataset was published in 2024 by Zimmer et al.~\cite{zimmer2024tumtraf} as part of the TUMTraf family of datasets. The dataset was collected at a large four-way intersection with one RSU and one CAV. The RSU was equipped with a 64-layer \SI{360}{\degree} LiDAR with \SI{120}{\metre} range and four 1920$\times$1200\,px RGB cameras providing a \SI{360}{\degree} surround view. The CAV was equipped with a 32-layer \SI{360}{\degree} LiDAR with a \SI{200}{\metre} measurement range and a 1920$\times$1200\,px RGB camera. Furthermore, the CAV was equipped with a GNSS sensor and an IMU. The dataset also provides HD map data for the covered area of the intersection. The dataset consists of 1,000 frames resulting in 2,000 labeled point clouds and 5,000 labeled images with a total of 30k 3D labels including track IDs of eight object classes including cars, trucks, pedestrians and cyclists. They used 800/100/100 frames for training, validation and testing.
The main task for this dataset is collective 3D object detection and tracking using camera and LiDAR-based methods. In particular, the dataset allows the evaluation of challenging scenes such as traffic violations and occlusions including VRUs. However, since the dataset contains only one CAV and one RSU, it is not suitable for V2V collective perception, nor can different V2X equipment rates be investigated. Furthermore, the low number of frames and the limited scenario diversity limit the applicability of the dataset.
\vspace*{0.2cm}
\subsubsection{\textbf{V2X-Real}}
The V2X-Real dataset was presented by Xiang et al. \cite{xiang2024v2x} in 2024. V2X-Real is a real-world dataset recorded at an intersection in Los Angeles. The dataset was recorded by two CAVs and two RSUs. The CAVs each were equipped with a 128-layer \SI{360}{\degree} LiDAR, four 1920$\times$1080 Stereo RGBD and a GPS. One of the RSUs was also equipped with a 128-layer \SI{360}{\degree} LiDAR, the other RSU was equipped with a 64-layer \SI{360}{\degree} LiDAR, in addition both RSUs contained two 1920$\times$1080 RGB cameras and a GPS each. The dataset consists of multiple scenarios at the RSU equipped intersection as well as V2V only scenarios at other intersections and roads. In total, this dataset consists of 33k labeled frames with 1.2M objects in the 10 classes car, pedestrian, scooter, motorcycle, bicycle, truck, van, barrier, box truck and bus. All sensors recorded at a frequency of \SI{10}{\hertz}, except the infrastructure cameras which recorded at \SI{30}{\hertz}. 
The main task for this dataset is collective 3D object detection, for which benchmarks are also provided for four subsets of the dataset. For the VC subset, the ego vehicle is fixed and the other vehicle as well as the infrastructure sensors are used as collaborative agents. For the IC subset, the infrastructure is selected as the ego and the other infrastructure sensors and the vehicles are the collaborative agents. In the V2V and I2I subsets, only vehicle and infrastructure data is used respectively.
This dataset is well suited to study V2X object detection, as it is the only real-world dataset containing multiple RSUs and CAVs, however the diversity of scenarios is limited.

\begin{table*}
\centering
\renewcommand{\arraystretch}{1.3}
\caption{Overview of V2V and V2X datasets for collective perception}
\resizebox{\textwidth}{!}{%
\begin{tabular}{lllllllllllll}
\hline
Dataset & Year & CAVs & RSUs & Frames & LiDAR & RGB Camera & Labels** & Classes & Scenarios & Data Source & Main Task & \begin{tabular}[c]{@{}l@{}}Directly \\ Available\end{tabular} \\ \hline
T\&J \cite{cooper}\begin{tabular}[c]{@{}l@{}}\ \\ \ \end{tabular} & 2019 & 2 & no & 100 & 16 Layer & - & - & ? & parking lot & Real World & Object Detection & \href{https://github.com/Aug583/F-COOPER}{\checkmark} \\ \hline
V2V-Sim \cite{wang2020v2vnet} \begin{tabular}[c]{@{}l@{}}\ \\ \ \end{tabular} & 2020 & $\sim$10 & no & 51k** & ? & - & ? & Car & ? & LiDARsim & Object Detection & \xmark \\ \hline
CODD \cite{codd}& 2021 & 4-16 & no & 13k & 64 Layer & - & 204k & Car, Pedestrian & \begin{tabular}[c]{@{}l@{}}urban, sub-urban, \\ rural\end{tabular} & CARLA & \begin{tabular}[c]{@{}l@{}}Object Detection,\\ Point Cloud Reg.\end{tabular} & \href{https://github.com/eduardohenriquearnold/CODD}{\checkmark} \\ \hline
COMAP \cite{comap} & 2021 & 2-20 & no & 8.6k & 32 Layer & 800$\times$600\,px* & 226k & Car, Truck & urban, sub-urban & CARLA, SUMO & \begin{tabular}[c]{@{}l@{}}Object Detection,\\ Sem. Seg.\end{tabular} & \href{https://github.com/YuanYunshuang/FPV\_RCNN}{\checkmark} \\ \hline
DAIR-V2X-C \cite{dair-v2x}& 2021 & 1 & yes & 13k & \begin{tabular}[c]{@{}l@{}}2x 300 Layer (RSU)\\ 40 Layer (CAV)\end{tabular} & \begin{tabular}[c]{@{}l@{}}2x 1920$\times$1080\,px (RSU)\\ 1920$\times$1080\,px (CAV)\end{tabular} & 464k & \begin{tabular}[c]{@{}l@{}}Vehicles, Motorcyclists, \\ Cyclists, Pedestrians\\ in 10 Classes\end{tabular} & urban & Real World & Object Detection & \xmark \\ \hline
V2XSet \cite{xu2022v2xvit} & 2022 & 2-7 & yes & 11k & 32 Layer & - & 254k & Car & \begin{tabular}[c]{@{}l@{}}urban, sub-urban, \\ rural, highway\end{tabular} & \begin{tabular}[c]{@{}l@{}}CARLA, OpenCDA, \\ SUMO\end{tabular} & Object Detection & \href{https://github.com/DerrickXuNu/v2x-vit}{\checkmark} \\ \hline
DOLPHINS \cite{dolphins} & 2022 & 2 & yes & 42k & 64 Layer (CAV) & \begin{tabular}[c]{@{}l@{}}1920$\times$1080\,px (RSU)\\ 1920$\times$1080\,px (CAV)\end{tabular} & 292k & Car, Pedestrian & \begin{tabular}[c]{@{}l@{}}urban, sub-urban, \\ rural, highway\end{tabular} & CARLA & Object Detection & \xmark \\ \hline
OPV2V \cite{opv2v} & 2022 & 2-7 & no & 11k & 64 Layer & 4x 800$\times$600\,px & 232k & Car & \begin{tabular}[c]{@{}l@{}}urban, sub-urban, \\ rural, highway\end{tabular} & \begin{tabular}[c]{@{}l@{}}OpenCDA,\\ CARLA, SUMO\end{tabular} & \begin{tabular}[c]{@{}l@{}}Object Detection\\ BEV Sem. Seg.\end{tabular} & \href{https://mobility-lab.seas.ucla.edu/opv2v/}{\checkmark} \\ \hline
V2X-Sim 2.0 \cite{V2XSim} & 2022 & 2-5 & yes & 10k & 32 Layer & \begin{tabular}[c]{@{}l@{}}4x 1600$\times$900\,px (RSU)\\ 6x 1600$\times$900\,px (CAV)\end{tabular} & 4.2M & \begin{tabular}[c]{@{}l@{}}Car, Cyclist, \\ Motorcyclist\end{tabular} & \begin{tabular}[c]{@{}l@{}}urban, sub-urban, \\ rural, highway\end{tabular} & CARLA, SUMO & \begin{tabular}[c]{@{}l@{}}Object Detection\\Sem. Seg.\\Tracking\end{tabular} & \href{https://ai4ce.github.io/V2X-Sim/index.html}{\checkmark} \\ \hline
IRV2V \cite{wei2023asynchrony} \begin{tabular}[c]{@{}l@{}}\ \\ \ \end{tabular}& 2023 & 2-5 & no & 8.4k & 32 Layer & 4x 600$\times$800\,px & 1.5M & ? & ? & CARLA &Object Detection & \xmark \\ \hline
DeepAccident \cite{wang2023deepaccident}& 2023 & 4 & yes & 57k & 32 Layer & \begin{tabular}[c]{@{}l@{}}6x 1600$\times$900\,px\\(CAV+RSU)\end{tabular}  & 2.1M & \begin{tabular}[c]{@{}l@{}}Vehicles, Motorcyclists, \\ Cyclists, Pedestrians\\ in 6 Classes\end{tabular} & \begin{tabular}[c]{@{}l@{}}urban, sub-urban, \\ rural, highway\end{tabular} & CARLA & Accident Prediction & \href{https://deepaccident.github.io/download.html}{\checkmark} \\ \hline
\begin{tabular}[c]{@{}l@{}}V2X Seq \cite{v2xseq}\\ Perception\end{tabular} & 2023 & 1 & yes & 15k & \begin{tabular}[c]{@{}l@{}}2x 300 Layer (RSU)\\ 40 Layer (CAV)\end{tabular} & \begin{tabular}[c]{@{}l@{}}2x 1920$\times$1080\,px (RSU)\\ 1920$\times$1080\,px (CAV)\end{tabular} & ? & \begin{tabular}[c]{@{}l@{}}Vehicles, Motorcyclists, \\ Cyclists, Pedestrians\\ in 10 Classes\end{tabular} & urban & Real World & \begin{tabular}[c]{@{}l@{}}Object Tracking\\ Object Detection\end{tabular} & \xmark \\ \hline
V2V4Real \cite{v2v4real} & 2023 & 2 & no & 9.7k & 32 Layer & \begin{tabular}[c]{@{}l@{}}1280$\times$720\,px (CAV1)*\\ 2064$\times$1544\,px (CAV2)*\end{tabular} & 240k & \begin{tabular}[c]{@{}l@{}}Vehicles, Pedestrians\\ in 5 Classes\end{tabular} & \begin{tabular}[c]{@{}l@{}}urban, sub-urban,\\ highway\end{tabular} & Real World & Object Detection & \href{https://mobility-lab.seas.ucla.edu/v2v4real/}{\checkmark} \\ \hline
LUCOOP \cite{lucoop}& 2023 & 3 & no & 7k & \begin{tabular}[c]{@{}l@{}}64 Layer (CAV1)\\ 16 Layer (CAV1)\\ 32 Layer (CAV2)\\ 16 Layer (CAV3)\end{tabular} & - & 34k & \begin{tabular}[c]{@{}l@{}}Vehicles, Cyclists,\\ Pedestrians\\ in 5 Classes\end{tabular} & urban & Real World & Object Detection & \href{https://data.uni-hannover.de/dataset/lucoop-leibniz-university-cooperative-perception-and-urban-navigation-dataset}{\checkmark} \\ \hline
V2X-Real \cite{xiang2024v2x}& 2024 & 2 & yes & 33k** & \begin{tabular}[c]{@{}l@{}}128 Layer (CAV)\\ 128 Layer (RSU1)\\ 64 Layer (RSU2)\end{tabular} & \begin{tabular}[c]{@{}l@{}}4x 1920$\times$1080\,px (CAV)\\ 2x 1920$\times$1080\,px (RSU)\end{tabular} & 1.2M & \begin{tabular}[c]{@{}l@{}}Vehicles, Motorcyclists, \\ Cyclists, Pedestrians\\ in 10 Classes\end{tabular} & urban & Real World & Object Detection & \xmark \\ \hline
TUMTraf-V2X \cite{zimmer2024tumtraf}& 2024 & 1 & yes & 1k & \begin{tabular}[c]{@{}l@{}}64 Layer (RSU)\\ 32 Layer (CAV)\end{tabular} & \begin{tabular}[c]{@{}l@{}}4x 1920$\times$1200\,px (RSU)\\ 1920$\times$1200\,px (CAV)\end{tabular} & 30k & \begin{tabular}[c]{@{}l@{}}Vehicles, Cyclists,\\ Pedestrians\\ in 8 Classes\end{tabular} & urban & Real World & \begin{tabular}[c]{@{}l@{}}Object Tracking\\ Object Detection\end{tabular} & \href{https://innovation-mobility.com/en/project-providentia/a9-dataset/}{\checkmark} \\ \hline
\end{tabular}%
}
\caption*{\small- not present in the dataset, * recorded, but currently not included, **may contain duplicates, ? unknown}

\label{tab:overview}
\end{table*}
\newpage
\section{DISCUSSION}
\label{sec:discussion}
\subsubsection{Object Detection}
The most prevalent task among the presented datasets is collective 3D object detection. While most datasets are well-suited for this task, there are significant differences between them.  None of the evaluated datasets clearly dominates the others. To study the impact of the connected automated vehicle (CAV) equipment rate on object detection capabilities, a wide range of CAV numbers is necessary. This applies to both the CODD and COMAP dataset. For this task, the CODD dataset is better suited due to its more diverse scenarios with more frames. However, both datasets only include LiDAR recordings, limiting the use of methods to LiDAR-based ones. 
For camera and LiDAR-camera methods the OPV2V dataset is well suited as it also features diverse scenarios. However, only cars are present in this dataset, so the detection of VRUs can't be evaluated. To evaluate VRU detection, the V2X-SIM dataset should be used, as it features the most diverse road users among the simulated datasets and also includes infrastructure sensors, allowing for the evaluation of V2I methods. However, the number of agents and diversity in scenarios is relatively low. The V2XSet dataset is outperformed by the other V2X datasets in most categories. However, it has the most diverse scenarios and the most agents among the V2X datasets. In terms of size, the DOLPHINS dataset clearly beats all other datasets and also features diverse scenarios and modalities. 
The number of agents in the real-world dataset is generally low, with the exception of the DAIR-V2X-C dataset, which has the highest number of agents. However, it is important to note that only one CAV is included in this dataset, with the other agents being RSUs. Therefore, only V2I methods can be evaluated. For V2V methods, LUCOOP or V2V4Real can be used. However, it is worth noting that V2V4Real only includes two CAVs, while LUCOOP has limited diversity in scenarios and does not provide a benchmark. The DAIR-V2X-C dataset is significantly larger than other real-world datasets, with more frames and labels. Also the TumTraf-V2X dataset can be used for object detection, however, the number of frames and collaborating agents is lower compared to the other presented real world datasets.
\subsubsection{Tracking \& Trajectory Prediction}
To evaluate tracking methods or trajectory prediction using a dataset, it is necessary to have knowledge of the trajectory of the moving objects. This can be achieved by assigning unique IDs to the labeled objects, which allows for tracking over consecutive frames. This information is available in V2X-Seq, DOLPHINS, OPV2V, V2X-Sim 2.0, and TumTraf-V2X. V2X-Seq and V2X-Sim are particularly suitable for this task due to the provided benchmarks. V2X-Seq is best suited for studying V2I tracking, while V2X-Sim is best suited for V2X tracking.
\subsubsection{Semantic Segmentaion}
To train and evaluate semantic segmentation methods, corresponding sensors require semantic labels. For example, cameras require per-pixel labels, while point clouds require per-point labels. In COMAP, semantic labels are provided for each sensor independently. In OPV2V, semantic labels are given as a BEV map for each CAV. V2X-SIM 2.0 is a versatile dataset for semantic segmentation, containing per sensor semantic labels and a semantic BEV map. However, COMAP has a larger number of agents, and OPV2V has a wider range of scenarios.
\subsubsection{Real World vs. Simulated Datasets}
To train and evaluate methods for real-world applications, real-world datasets are preferable due to the sim2real gap of simulated datasets. However, acquiring these datasets is costly and time-intensive. As a result, there are only a few available datasets with limitations regarding the number of CAVs, frames, and modalities. Therefore, it makes sense to use simulated datasets for training and evaluation, especially if the method is not directly applied to real-world applications. However, to obtain expressive results, the simulation must closely resemble reality. All publicly available datasets use CARLA as a simulator, which is well-suited for this task. However, the LiDAR model, in particular, has some fidelity disadvantages. 
Although weather can be parameterized in CARLA, the LiDAR sensor is not affected by precipitation or fog. Interference can only be simulated by increasing the drop-off rate and detection noise, where the drop-off is not distance dependent and the noise is generated by shifting the detection by a random distance along the ray. However, any reference to a real measurement under adverse weather conditions is missing. Furthermore, the sensor only returns an intensity, which is calculated by the atmospheric attenuation factor and the distance to the object. Other variables that influence the intensity of the reflected signal, such as material properties of the hit object and angle of incidence of the light beam, are not taken into account. Moreover, physical properties such as beam divergence or a detection threshold are missing. So this large domain gap should be considered when choosing a simulated dataset.
\section{CONCLUSION}
\label{sec:conclusion}

This paper presents a comprehensive review of collective perception datasets for autonomous driving. We provide recommendations on the most suitable dataset for collective 3D object detection, tracking and semantic segmentation based on acquired knowledge. The review focuses on V2V and V2X datasets, including both synthetic and real-world datasets capable of V2V or V2X collective perception. Additionally, various infrastructure-sensor datasets suitable for collective perception are considered. We analyzed several critical characteristics of the datasets, including the number of frames, number of collaborative agents, diversity in scenarios and object classes, and public availability. For publicly available datasets, we conducted an in-depth analysis to investigate potential anomalies.

\addtolength{\textheight}{-2cm}   
\bibliographystyle{IEEEtran} 
\bibliography{literature.bib}

\end{document}